\documentclass{article}

\usepackage[preprint,nonatbib]{neurips_2025}

\usepackage[utf8]{inputenc} 
\usepackage[T1]{fontenc}    
\usepackage{hyperref}       
\usepackage{url}            
\usepackage{booktabs}       
\usepackage{tabularx}
\usepackage{amsfonts}       
\usepackage{nicefrac}       
\usepackage{microtype}      
\usepackage{xcolor}         
\usepackage{amsmath}
\usepackage{xcolor}
\usepackage{graphicx}
\usepackage{multicol}
\usepackage{multirow}
\usepackage[ruled]{algorithm2e}
\usepackage{colortbl} 
\usepackage{xcolor}   
\usepackage{wrapfig}
\usepackage{graphicx}
\usepackage{caption}
\usepackage{enumitem}

\definecolor{lightpurple}{rgb}{0.9, 0.9, 1.0} 

\title{Reinforced Internal-External Knowledge Synergistic Reasoning for Efficient Adaptive Search Agent}

\author{
Ziyang Huang\textsuperscript{$\alpha\beta$}, Xiaowei Yuan\textsuperscript{$\alpha\beta\gamma$}, Yiming Ju\textsuperscript{$\gamma$}, Jun Zhao\textsuperscript{$\alpha\beta$}, Kang Liu\textsuperscript{$\alpha\beta$} \\
\textsuperscript{$\alpha$}Institute of Automation, Chinese Academy of Sciences \\
\textsuperscript{$\beta$}University of Chinese Academy of Sciences \\
\textsuperscript{$\gamma$}Beijing Academy of Artificial Intelligence \\
\texttt{huangziyang2023@ia.ac.cn} \\
\textbf{Code}: \url{https://github.com/hzy312/knowledge-r1}
}

\begin{document}

\maketitle

\begin{abstract}

Retrieval-augmented generation (RAG) is a common strategy to reduce hallucinations in Large Language Models (LLMs). While reinforcement learning (RL) can enable LLMs to act as search agents by activating retrieval capabilities, existing ones often underutilize their internal knowledge.  This might lead to redundant retrievals, potential harmful knowledge conflicts, and increased inference latency. To address these limitations, an efficient adaptive search agent capable of discerning optimal retrieval timing and synergistically integrating parametric (internal) and retrieved (external) knowledge is in urgent need. This paper introduces the Reinforced \textbf{I}nternal-External \textbf{K}nowledge Synergistic R\textbf{E}asoning \textbf{A}gent (\textbf{IKEA}), which could indentify its own knowledge boundary and prioritize the utilization of internal knowledge, resorting to external search only when internal knowledge is deemed insufficient. This is achieved using a novel knowledge-boundary aware reward function and a knowledge-boundary aware training dataset. These are designed for internal-external knowledge synergy oriented RL, incentivizing the model to deliver accurate answers, minimize unnecessary retrievals, and encourage appropriate retrievals when its own knowledge is lacking. Evaluations across multiple knowledge-intensive reasoning tasks demonstrate that IKEA significantly outperforms baseline methods, reduces retrieval frequency significantly, and exhibits robust generalization capabilities.

\end{abstract}

\section{Introduction}

The advancement of large-scale reinforcement learning (RL) with verifiable reward systems \cite{shao2024deepseekmathpushinglimitsmathematical,su2025crossingrewardbridgeexpanding} has significantly enhanced the capabilities of reasoning models like Deepseek R1 \cite{deepseekai2025deepseekr1incentivizingreasoningcapability}. For knowledge-intensive tasks \cite{gao2024retrievalaugmentedgenerationlargelanguage}, R1-like models could activate their internal pre-trained knowledge through reasoning. However, constrained by the finite nature of pre-training corpora and the dynamic essence of world knowledge, they remain susceptible to hallucinations \cite{Huang_2025}. To address the knowledge deficiencies, current research typically empowers models to invoke search engines, essentially training them as search agents \cite{jin2025searchr1trainingllmsreason, song2025r1searcherincentivizingsearchcapability, chen2025researchlearningreasonsearch}. During reinforcement learning, these models progressively learn to decompose tasks and retrieve relevant knowledge for each subtask to aid reasoning. Despite this, the approach remains suboptimal for several reasons:

Firstly, it primarily leverages the tool-calling and information extraction capabilities of LLMs, largely underutilizing its potential as an intrinsic knowledge base (i.e., LLM-as-KB) \cite{heinzerling-inui-2021-language, zheng2024reliablellmsknowledgebases}. This leads to substantial retrieval redundancy, as external searches are performed even when the necessary information might already be implicitly encoded within the parameters of the model. Secondly, the limited capabilities of the retriever can introduce noise into the retrieval results \cite{10.1145/3696410.3714717}, potentially generating unnecessary knowledge conflicts \cite{fang-etal-2024-enhancing}. A common issue is erroneous retrieved knowledge overriding the accurate parametric knowledge \cite{xu-etal-2024-knowledge-conflicts}. Furthermore, the iterative nature of search engine calls frequently interrupts the generation process of LLM, resulting in significant inference delays \cite{yu2024autoragautonomousretrievalaugmentedgeneration}. Thus, a critical research question emerges: \textit{\textbf{How can we train an efficient adaptive search agent that comprehensively integrates both parametric (internal) and retrieved (external) knowledge?}}

This paper argues that such an agent needs to posses the following three key knowledge behaviors: (1) Self-knowledge Boundary Division (determine know/unknown): the ability to decompose a query into atomic queries and determine whether each sub-query falls within the knowledge boundary of the agent \cite{li2024knowledgeboundarylargelanguage, ren2024investigatingfactualknowledgeboundary, wen2024perception}; (2) Internal Knowledge Recall (search in parameter): the ability to generate relevant background knowledge to assist in answering questions that fall within its knowledge boundary \cite{cheng2024understandinginterplayparametriccontextual, mao-etal-2021-generation}; (3) External Knowledge Recall (search in corpus): the ability to generate effective search queries for questions outside its knowledge boundary and utilize search engines to acquire the desired knowledge \cite{zhao2024knowingllmsknowsimple}. 
In all, an efficient adaptive search agent needs to accurately determine whether to search in parameter or corpus, and should minimize the use of external knowledge by leveraging its internal knowledge as much as possible. Therefore, the retrieval timing becomes the core. Existing research determines retrieval timing either via external indicators/classifiers, which often generalize poorly and require external tools \cite{jiang2023activeretrievalaugmentedgeneration, jeong2024adaptiveraglearningadaptretrievalaugmented}, or through complex data engineering for imitation/preference learning to enable autonomous decision-making \cite{yu2024autoragautonomousretrievalaugmentedgeneration, guan2025deepragthinkingretrievalstep, wang2025chainofretrievalaugmentedgeneration}. However, how to imbue a model with the capacity to determine the optimal retrieval timing for adaptive retrieval via RL has not been fully investigated.

To address these issues and enable the model to exhibit the aforementioned knowledge behaviors, we propose the Reinforced \textbf{I}nternal-External \textbf{K}nowledge Synergistic R\textbf{E}asoning \textbf{A}gent (\textbf{IKEA}), an efficient adaptive search agent powered by RL \cite{schulman2017proximalpolicyoptimizationalgorithms, shao2024deepseekmathpushinglimitsmathematical}. First, we design the IKEA agent framework, which explicitly prompts the model to determine its internal knowledge boundary and prioritize the utilization of knowledge within its parameters. The search engine is invoked to retrieve external knowledge only when internal knowledge is deemed uncertain or insufficient. Next, we introduce two key components: a knowledge-boundary aware reward function and a corresponding knowledge-boundary aware training dataset for internal-external knowledge synergy oriented RL. The reward function incentivizes correct answers while minimizing unnecessary external knowledge retrieval for questions where the LLM possesses sufficient internal knowledge, and conversely, encouraging retrieval for questions beyond its internal knowledge boundary. This approach aims to improve the perception of its self-knowledge. The training dataset, meticulously constructed, comprises an equal mix of questions that the model is likely to answer using its internal knowledge and those requiring external knowledge. This balanced dataset is crucial for training the model to adaptively and synergistically leverage both internal and external knowledge.

We conducted evaluations on multiple datasets involving both single-hop \cite{kwiatkowski-etal-2019-natural, mallen-etal-2023-trust} and multi-hop \cite{yang-etal-2018-hotpotqa, ho-etal-2020-constructing} knowledge reasoning tasks. IKEA outperforms baseline methods across various datasets, achieving exceptional performance, and demonstrates strong generalization capabilities on out-of-distribution datasets. Compared to naive reinforcement learning approaches (i.e. Search-R1) \cite{jin2025searchr1trainingllmsreason, song2025r1searcherincentivizingsearchcapability, chen2025researchlearningreasonsearch}, it can significantly reduce the number of retrievals while improving performance. This fully showcases the effectiveness and efficiency of our proposed method. The contribution of this paper are as follows:

\begin{itemize}[leftmargin=*, nolistsep]
\setlength{\itemsep}{1mm}
    \item This paper addresses the limitations of current search agents, which often over-rely on external searches and underutilize their intrinsic knowledge, leading to retrieval redundancy.
    \item This paper proposes Reinforced Internal-External Knowledge Synergistic Reasoning Agent (IKEA), an efficient adaptive search agent via reinforcement learning, which could delineate the self-knowledge boundary and prioritize parametric knowledge before resorting to external retrieval.
    \item This paper provides detailed analysis to explain that knowledge-boundary aware reward design and data construction are both key to training efficient adaptive search agents.
\end{itemize}

\section{Preliminary}
\label{sec:rl}

\subsection{Multi-turn Reinforcement Learning with Verifiable Reward for Large Language Model} 
This work considers an LLM agent $\pi$ operating within an environment $E$ to complete a task $t$. The interaction proceeds in $N$ rounds: in each round $k$, the agent takes action $a_k$, receives observation $o_{k+1}$. Both actions and observations are token sequences: $a_k = (a_{k,1}, \dots, a_{k,\ell}), o_{k+1}=(o_{k+1,1}, \dots, o_{k+1, \ell})$. The state $s_k$ at round $k$ is the concatenation of all preceding tokens $(t, a_0, o_1, \dots, a_{k-1}, o_k)$. Upon task completion after $N$ rounds, a final reward $r$ is provided by the reward model. A trajectory is like: $\tau = (t, a_0, o_1, \dots, a_{N-1}, o_N, r)$.
Reinforcement Learning trains the policy $\pi(a|s)$ on collected trajectories to maximize the expected cumulative reward, aiming for an optimal policy that maximizes the total reward.

Proximal Policy Optimization (PPO) \cite{schulman2017proximalpolicyoptimizationalgorithms} is a common RL baseline, optimizing the policy via the clipped surrogate loss:

{
\small
\begin{align}
    \mathcal{L}^{\textrm{PPO}}(\theta) = -\hat{\mathbb{E}}_{t \sim T}  \frac{1}{\sum_{k=0}^{N-1} |a_k|}  \sum_{k=0}^{N-1} \sum_{\ell=1}^{|a_k|}  \left[\min\left( r_\theta \hat{A}_{\tau}, \text{clip}(r_\theta, 1-\epsilon, 1+\epsilon) \hat{A}_{\tau}\right)\right],
     r_\theta = \frac{\pi_\theta(a_{k,l}|s_k)}{\pi_{\theta_{old}}(a_{k,l}|s_k)}
\end{align}
}

$\hat{A}_{\tau}$ is the advantage, and $\epsilon$ is the clipping ratio. It is noteworthy that we only compute the loss on the action tokens. We mask the loss from the observation tokens because they are from the external environment and are not generated by the LLM. However, PPO requires training a separate value model (typically similar in size to the policy model) to estimate the value function $V$ and compute the advantage $\hat{A}$ (often using GAE \cite{schulman2018highdimensionalcontinuouscontrolusing}), resulting in significant additional memory overhead.

Therefore, this paper adopts Group Relative Policy Optimization (GRPO) \cite{shao2024deepseekmathpushinglimitsmathematical} as the default RL algorithm. As shown in the top of the Figure \ref{fig:method}, GRPO performs multiple rollouts per task and calculate the relative reward within the group as the advantage. This method avoids the need for a separate value model and has shown performance comparable to or exceeding PPO. Its loss function is:
{
\small
\begin{align}
    &\mathcal{L}^{\textrm{GRPO}}(\theta) = -\hat{\mathbb{E}}_{t \sim T, \tau_i \sim \pi_{old}(\tau|t)}  \frac{1}{G}  \sum_{i=1}^G \frac{1}{\sum_{k=0}^{N-1} |a_{i,k}|} \nonumber \\
    &\sum_{k=0}^{N-1} \sum_{\ell=1}^{|a_{i,k}|}   \left[\min\left( r_\theta \hat{A}_{\tau_i}, \text{clip}(r_\theta, 1-\epsilon, 1+\epsilon) \hat{A}_{\tau_i}\right) - \beta \mathbb{D}_{KL}[\pi_{\theta} || \pi_{old}]\right], r_\theta = \frac{\pi_\theta(a_{i,k,l}|s_{i,k})}{\pi_{\theta_{old}}(a_{i,k,l}|s_{i,k})} 
\end{align}
}

where $\hat{A}_{\tau_i} = \frac{r_i - \mu_r}{\sigma_r}$ is the estimated advantage of trajectory $\tau_{i}$ based on group-relative rewards. $\mu_r$  is the mean and the $\sigma_r$ is the standard deviation of the rewards within the group.

\subsection{Knowledge Boundary of Large Language Model}
We use the term \textit{Knowledge Boundary} to distinguish between the internal and external knowledge of a specific LLM. Internal knowledge refers to the knowledge that can be extracted from the model through some knowledge probing method, while external knowledge refers to the relative complement of the internal knowledge in the whole world knowledge; that is, the knowledge that does not exist within the parameters of the model. We also use knowledge boundary to differentiate between various questions. Whether a question falls inside or outside the knowledge boundary refers to whether the knowledge required to answer that question is internal knowledge or external knowledge.
\section{Method}
\begin{figure}
    \centering
    \includegraphics[width=\linewidth]{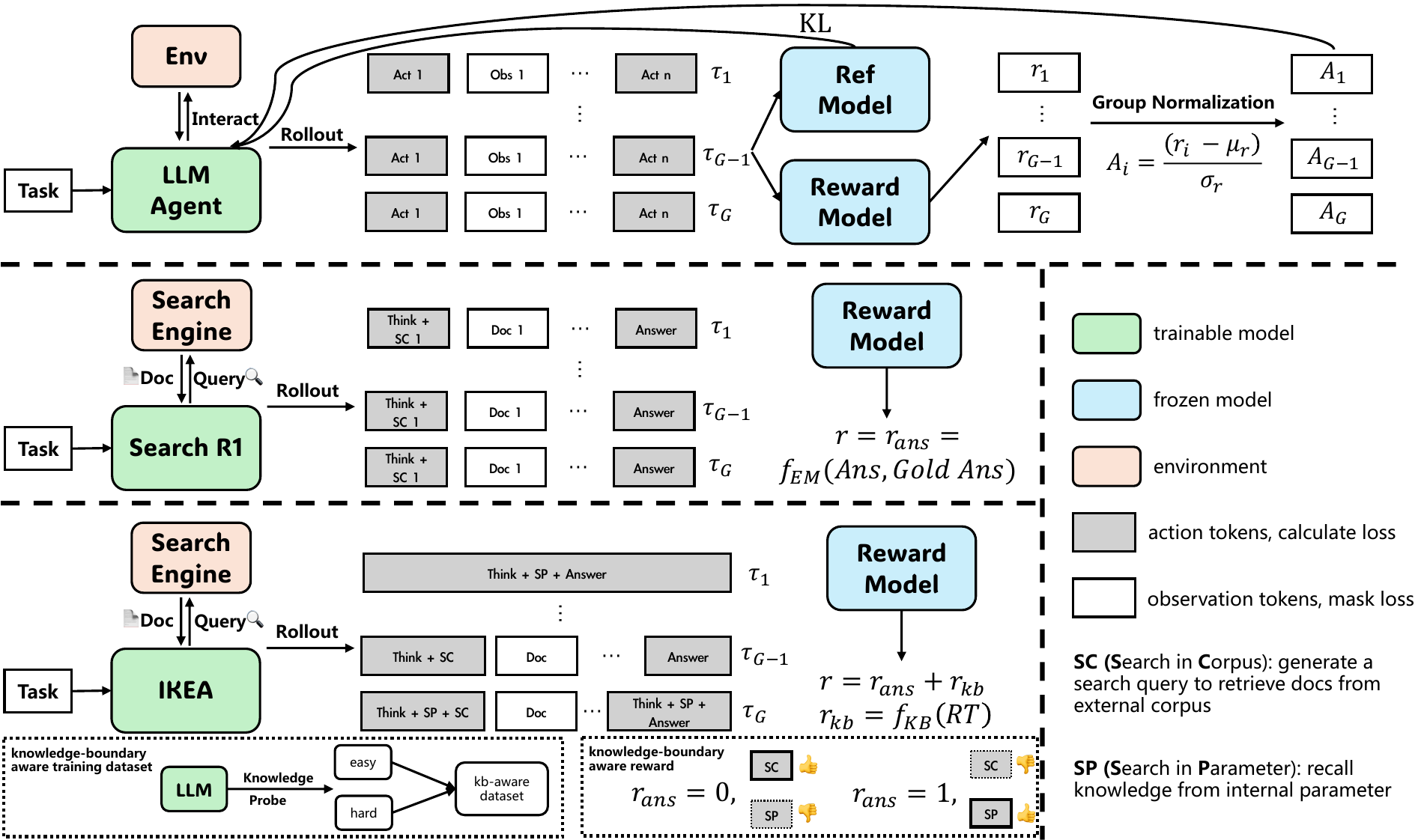}
    \caption{The top of the figure illustrates the training process for Multi-turn Reinforcement Learning with Verifiable Reward for LLM-Agent. In the middle is Search-R1, and at the very bottom is IKEA. Search-R1 and IKEA are special types of LLM-agents. We highlight the differences from the training of general LLM-agents, and to save space, we have omitted the common parts, such as the calculation of KL and Advantage.}
    \label{fig:method}
\end{figure}

\subsection{Basic Setting}
\label{sec:searchr1}
In the context of knowledge-intensive reasoning, each task is framed as a query related to world knowledge. The environment (search engine) comprises a text corpus and a retriever. The agent interacts with the environment by generating a sequence of action tokens and receiving a sequence of observation tokens. As illustrated in the middle of the Figure \ref{fig:method}, a typical LLM-based search agent \cite{jin2025searchr1trainingllmsreason, song2025r1searcherincentivizingsearchcapability, chen2025researchlearningreasonsearch} will generate reasoning thought, search query, and final answer in its action tokens. We detail the agentic workflow and training method of this line of work as follows:

To facilitate the parsing of the executable actions for interaction with the environment, we define three sets of special tags to structure the action token sequence: \textsc{<think>[reasoning content]</think>}, \textsc{<search>[search query]</search>}, and \textsc{<answer>[final answer]</answer>}.
It is noteworthy that although the content within the \textsc{<think>} tags does not directly interact with the environment, it is generated by the model and is considered part of the action token sequence. In each turn, the agent must first generate reasoning content within the \textsc{<think>} tag to analyze the current state and subsequently generate either a \textsc{<search>} or an \textsc{<answer>} tag to interact. When a \textsc{<search>} tag is generated, the model produces a search query within it, which is then used by the retriever to get relevant knowledge from the corpus.
We define a special set of tags: \textsc{<context>[retrieved context]</context>}, and the retrieved relevant documents are placed between these tags and inserted as an observation after the generated token sequence, allowing the interaction to continue to the next turn. The content within the \textsc{<context>} tags is not generated by the model and is therefore masked out during loss calculation. When an \textsc{<answer>} tag is generated, the model outputs the answer of the task query, at which point the entire task execution process completes. We refer to such a complete process as a rollout. Upon obtaining the final answer, we evaluate the reward for the current trajectory using an Exact Match Reward Function. Then we can repeat the rollout processing many times to get a group of trajectories and leverage the loss function detailed in Section \ref{sec:rl} to optimize the agent.

\subsection{IEKA: Reinforced Internal-External Knowledge Synergistic Reasoning Agent}
\label{method:ikea}

Existing search agents often primarily leverage the planning capability of the LLM, decomposing a task query into multiple subqueries and iteratively retrieving relevant evidence documents for each subquery to aid reasoning. Such search agents fail to fully utilize the capacity of the LLM as a parametric knowledge base, resulting in much redundant retrieval. This not only introduces significant inference latency but also potentially leads to harmful knowledge conflicts \cite{xu-etal-2024-knowledge-conflicts, jin-etal-2024-cutting} (wrong external knowledge overrides the correct internal knowledge).  Building upon this, this paper argues that an efficient adaptive search agent is needed to address these problems. It should possess the ability to delineate its own knowledge boundary and leverage internal parametric knowledge as much as possible within this boundary, while employing retrieval for knowledge outside this boundary.

To this end, we propose the Reinforced \textbf{I}nternal-External \textbf{K}nowledge Synergistic R\textbf{E}asoning \textbf{A}gent (\textbf{IKEA}). As shown in the bottom of Figure \ref{fig:method}, this paper first designs the agent prompt template to enable the model to autonomously perform synergistic reasoning using both internal and external knowledge. Subsequently, we design a knowledge-boundary aware reward function and construct a knowledge-boundary aware training dataset, which will encourage the model to clarify its own knowledge boundary while utilizing different knowledge behaviors both inside and outside this boundary. Finally, we apply reinforcement learning to finetune the agent towards internal-external knowledge synergy.

\paragraph{IKEA Agent Prompt Template}
To begin, we incorporate output format constraints into the prompt to ensure the agent interacts with the environment in the format described in the Section \ref{sec:searchr1}. Furthermore, we prompt the model to evaluate all subqueries, encouraging it to utilize its internal parametric knowledge whenever it is confident. When the model is uncertain about the knowledge regarding specific information, it is encouraged to retrieve relevant knowledge from the external knowledge base. The detailed prompt template is provided in the Appendix \ref{app:prompt}.

\paragraph{Knowledge-boundary Aware Reward Design}

Due to the probabilistic nature of LLMs, existing LLMs have a blurred perception of their self-knowledge boundaries. They cannot definitively distinguish which questions pertain to internal knowledge and which require external knowledge. As shown in the bottom of the Figure \ref{fig:method}, for the same task, $\tau_1$ only uses internal knowledge, $\tau_{G-1}$ only use external knowledge, and $\tau_{G}$ uses both internal and external knowledge. Consequently, prompt-based agents may exhibit knowledge misidentification behaviors, leading to the generation of hallucinated answers for questions outside their knowledge boundaries, while utilizing redundant retrieval to for questions within their knowledge boundaries.

To address this, we design a knowledge-boundary aware reward composed of several components. First, the answer reward ($r_{ans}$) is 1 if the final answer matches the gold answer, and 0 otherwise. Second, the knowledge boundary reward ($r_{kb}$) is determined as follows: if $r_{ans} = 1$, $r_{kb}$ is a linear function increasing as the retrieval times ($RT$) decrease, ranging from 0 to $r_{kb+}$. If $r_{ans} = 0$, then $r_{kb} = 0$ when the number of retrievals is 0, and $r_{kb} = r_{kb-}$ (a small value) when the number of retrievals is greater than 0. Finally, for the format reward, if the generated trajectory violates format constraints of IKEA, the total reward is -1; otherwise, it is $r_{ans} + r_{kb}$.

The expression for the reward function is as follows:

\begin{align}
&R =
\begin{cases}
-1 & \text{if trajectory format is incorrect} \\
r_{\text{ans}} + r_{kb} & \text{if trajectory format is correct}
\end{cases}
\\
&r_{\text{ans}} =
\begin{cases}
1 & \text{ans == gold ans} \\
0 & \text{ans != gold ans}
\end{cases},
r_{kb} =
\begin{cases}
r_{kb+} \times \left(1 - \frac{RT}{RT_{max}}\right) & \text{if } r_{\text{ans}}=1 \\
0 & \text{if } r_{\text{ans}}=0 \text{ and } RT=0 \\
r_{kb-} & \text{if } r_{\text{ans}}=0 \text{ and } RT>0
\end{cases}
\end{align}

Here, $RT_{max}$ denotes the maximum number of retrievals, $r_{kb-}$ is a small value, $r_{kb+}$ is the maximum possible knowledge boundary reward. During exploration, when the agent obtains the correct answer ($r_{ans} = 1$), it may utilize internal or external knowledge. The reward $r_{kb+}$ is designed to incentivize the agent to minimize retrieval attempts, thereby favoring the use of internal knowledge. Conversely, when the agent fails to obtain the correct answer ($r_{ans} = 0$), indicating high uncertainty regarding relevant knowledge, the reward $r_{kb-}$ encourages reliance on external knowledge. To prevent the development of excessive retrieval behavior, we establish $r_{kb-} \ll r_{kb+}$.

\paragraph{Knowledge-boundary Aware Dataset Construction}
We use In-context Learning with three Chain-of-Thought exemplars to probe the internal knowledge of the model. For each question, we sample the answer $N$ times. A question is labeled $Q_{easy}$ if the correct answer is obtained at least once, indicating the model likely possesses the relevant knowledge. Otherwise, it's labeled $Q_{hard}$.

If the training dataset exclusively contained data from $Q_{easy}$, the model would be more likely to utilize internal knowledge during rollout, and relying solely on internal knowledge would yield higher rewards than using retrieval. Consequently, after full training, the model would tend to avoid retrieval for any question. Conversely, if the training dataset only comprised $Q_{hard}$ questions, the model would be more inclined to use external retrieved knowledge during the rollout, and using the retriever would result in higher rewards than not using it. Thus, after full training, the model would tend to use retrieval exclusively for all questions.

To achieve a balanced use of internal and external knowledge, we construct the training dataset with a 1:1 ratio of $Q_{easy}$ and $Q_{hard}$ questions. This promotes adaptive retrieval and synergy between internal and external knowledge.

\section{Experiment}

\subsection{Setting}


Test sets (easy and hard subsets) were constructed like the training set (Section \ref{method:ikea}), including two in-distribution and two out-of-distribution sets (details in Appendix \ref{app:dataset}). We benchmarked our method against baselines (Appendix \ref{app:baseline}) using various model sizes and types, with training specifics in Appendix \ref{app:implementation}. Performance was evaluated by exact match (EM) and efficiency by the number of valid searches (RT).

\begin{table*}[t!]
\centering
%
\caption{Performance of Qwen2.5-3B and Qwen2.5-7B (Base \& Instruct). "-Zero" are trained from Base. EM = exact match, RT = number of valid searches. The original checkpoint of Search-R1-Zero-3B might be over-optimized (hard to count RT). \textsuperscript{\dag}DeepRAG EM/RT results are from its paper.}
\label{tab:main_table}
\resizebox{\textwidth}{!}{%
\begin{tabular}{@{}lcccccccccccccccccc@{}}
\toprule
\multirow{3}{*}{\textbf{Method}} & \multicolumn{4}{c}{\textbf{NQ}} & \multicolumn{4}{c}{\textbf{PopQA}} & \multicolumn{4}{c}{\textbf{HotpotQA}} & \multicolumn{4}{c}{\textbf{2Wiki}} & \multicolumn{2}{c}{\multirow{2}{*}{\textbf{Avg}}} \\
\cmidrule(lr){2-5} \cmidrule(lr){6-9} \cmidrule(lr){10-13} \cmidrule(lr){14-17} 
& \multicolumn{2}{c}{Easy} & \multicolumn{2}{c}{Hard} & \multicolumn{2}{c}{Easy} & \multicolumn{2}{c}{Hard} & \multicolumn{2}{c}{Easy} & \multicolumn{2}{c}{Hard} & \multicolumn{2}{c}{Easy} & \multicolumn{2}{c}{Hard} & \multicolumn{2}{c}{} \\ 
\cmidrule(lr){2-3} \cmidrule(lr){4-5} \cmidrule(lr){6-7} \cmidrule(lr){8-9} \cmidrule(lr){10-11} \cmidrule(lr){12-13} \cmidrule(lr){14-15} \cmidrule(lr){16-17} \cmidrule(lr){18-19} 
& EM & RT & EM & RT & EM & RT & EM & RT & EM & RT & EM & RT & EM & RT & EM & RT & EM & RT \\
\midrule
\multicolumn{19}{@{}l}{\textbf{Qwen2.5-3B}} \\
\multicolumn{19}{@{}l}{\textit{w/o parameter update} (re-implementation)} \\
Direct & 36.33 & 0 & 3.91 & 0 & 56.05 & 0 & 2.54 & 0 & 50.39 & 0 & 1.56 & 0 & 50.98 & 0 & 11.72 & 0 & 26.69 & 0.00 \\
RAG & 59.77 & 1 & 30.47 & 1 & 68.16 & 1 & 31.64 & 1 & 54.30 & 1 & 13.87 & 1 & 40.04 & 1 & 12.70 & 1 & 38.87 & 1.00 \\
Iter-Retgen & 59.57 & 4 & 30.27 & 4 & 68.55 & 4 & 32.81 & 4 & 55.86 & 4 & 16.02 & 4 & 41.60 & 4 & 15.23 & 4 & 39.99 & 4.00 \\
IR-COT & 35.74 & 3.26 & 15.04 & 3.34 & 48.05 & 3.15 & 24.80 & 3.17 & 43.36 & 3.64 & 9.77 & 3.60 & 25.39 & 3.82 & 9.18 & 3.70 & 26.42 & 3.46 \\
FLARE & 34.57 & 0.21 & 4.49 & 0.41 & 52.15 & 0.18 & 4.49 & 0.52 & 48.44 & 0.16 & 1.76 & 0.61 & 50.00 & 0.03 & 11.13 & 0.32 & 25.88 & 0.31 \\
\midrule
\multicolumn{19}{@{}l}{\textit{Reinforcement learning} (re-implementation for search-r1)} \\
R1-Zero & 62.34 & 0 & 10.55 & 0 & 72.66 & 0 & 3.71 & 0 & 59.57 & 0 & 4.49 & 0 & 57.22 & 0 & 13.28 & 0 & 35.48 & 0.00 \\
R1 & 59.77 & 0 & 7.23 & 0 & 70.11 & 0 & 3.13 & 0 & 58.01 & 0 & 3.71 & 0 & 57.81 & 0 & 13.67 & 0 & 34.18 & 0.00 \\
Search-R1-Zero\textsuperscript{***} & 66.60 & - & 28.51 & - & 77.73 & - & 27.93 & - & 64.45 & - & 13.67 & - & 52.54 & - & 13.48 & - & 43.11 & - \\
Search-R1 & 66.41 & 1.17 & 32.61 & 1.30 & 73.43 & 1.22 & 29.49 & 1.53 & 65.23 & 1.86 & 22.27 & 1.88 & 51.17 & 2.16 & 26.56 & 2.00 & 45.90 & 1.64 \\
\rowcolor{lightpurple} \textbf{IKEA-Zero} & 71.29 & 1.00 & 34.18 & 1.00 & 78.90 & 1.00 & 35.94 & 1.02 & 68.94 & 1.05 & 21.09 & 1.14 & 54.69 & 1.19 & 23.63 & 1.39 & \textbf{48.58 (+5.47)} & \textbf{1.10} \\
\rowcolor{lightpurple} \textbf{IKEA} & 72.46 & 1.00 & 31.44 & 1.02 & 79.69 & 1.00 & 33.59 & 1.02 & 69.92 & 1.04 & 20.11 & 1.13 & 59.37 & 1.15 & 20.70 & 1.21 & \textbf{48.41 (+2.51)} & \textbf{1.07 (-34.76\%)} \\
\midrule\midrule 
\multicolumn{19}{@{}l}{\textbf{Qwen2.5-7B}} \\
\multicolumn{19}{@{}l}{\textit{w/o parameter update} (re-implementation)} \\
Direct & 41.41 & 0 & 4.30 & 0 & 61.13 & 0 & 2.34 & 0 & 54.69 & 0 & 4.10 & 0 & 51.95 & 0 & 10.94 & 0 & 28.86 & 0.00 \\
RAG & 57.23 & 1 & 26.37 & 1 & 69.73 & 1 & 31.64 & 1 & 58.98 & 1 & 17.77 & 1 & 40.82 & 1 & 11.33 & 1 & 39.23 & 1.00 \\
Iter-Retgen & 58.79 & 4 & 26.95 & 4 & 70.90 & 4 & 31.25 & 4 & 61.52 & 4 & 19.73 & 4 & 43.36 & 4 & 14.45 & 4 & 40.87 & 4.00 \\
IR-COT & 40.04 & 2.59 & 14.26 & 2.68 & 58.98 & 2.48 & 25.78 & 2.56 & 46.09 & 3.09 & 14.26 & 2.94 & 17.58 & 3.18 & 12.50 & 3.07 & 28.69 & 2.82 \\
FLARE & 39.65 & 0.16 & 5.27 & 0.28 & 59.96 & 0.11 & 3.13 & 0.67 & 52.93 & 0.08 & 4.10 & 0.375 & 51.17 & 0.03 & 11.52 & 0.35 & 28.47 & 0.26 \\
\midrule
\multicolumn{19}{@{}l}{\textit{SFT/DPO} (results from the original paper, shown as EM/RT)} \\
DeepRAG\textsuperscript{\dag} & \multicolumn{4}{c}{-} & \multicolumn{4}{c}{40.60/} & \multicolumn{4}{c}{32.10/} & \multicolumn{4}{c}{40.40/} & - & -\\
\midrule
\multicolumn{19}{@{}l}{\textit{Reinforcement learning}} \\
R1-Zero & 66.80 & 0 & 15.23 & 0 & 72.65 & 0 & 6.25 & 0 & 64.65 & 0 & 5.66 & 0 & 53.32 & 0 & 18.16 & 0 & 37.84 & 0.00 \\
R1 & 62.50 & 0 & 14.06 & 0 & 73.04 & 0 & 5.27 & 0 & 64.06 & 0 & 5.47 & 0 & 57.23 & 0 & 14.45 & 0 & 37.01 & 0.00 \\
Search-R1-Zero & 68.55 & 1.19 & 35.55 & 1.34 & 76.37 & 1.16 & 33.59 & 1.30 & 69.73 & 1.78 & 25.78 & 1.77 & 46.68 & 2.38 & 26.56 & 2.13 & 47.85 & 1.63 \\
Search-R1 & 65.63 & 1.34 & 33.40 & 1.51 & 78.13 & 1.24 & 32.62 & 1.51 & 68.17 & 2.00 & 24.02 & 2.07 & 35.35 & 2.67 & 22.66 & 2.47 & 45.00 & 1.85 \\
\rowcolor{lightpurple} \textbf{IKEA-Zero} & 74.80 & 1.00 & 37.89 & 1.00 & 80.47 & 1.00 & 33.20 & 1.00 & 74.22 & 1.01 & 23.43 & 1.08 & 57.42 & 1.03 & 27.34 & 1.23 & \textbf{51.10 (+3.25)} & \textbf{1.04 (-36.20\%)} \\
\rowcolor{lightpurple} \textbf{IKEA} & 74.61 & 0.59 & 32.23 & 0.89 & 80.08 & 0.56 & 31.84 & 1.09 & 71.88 & 0.60 & 26.56 & 1.20 & 54.49 & 0.93 & 28.71 & 1.38 & \textbf{50.05 (+5.05)} & \textbf{0.91 (-50.81\%)} \\
\bottomrule
\end{tabular}
} 
\end{table*}
\subsection{Overall Results}


\begin{figure*}[t!]
    \centering
    \includegraphics[width=\linewidth]{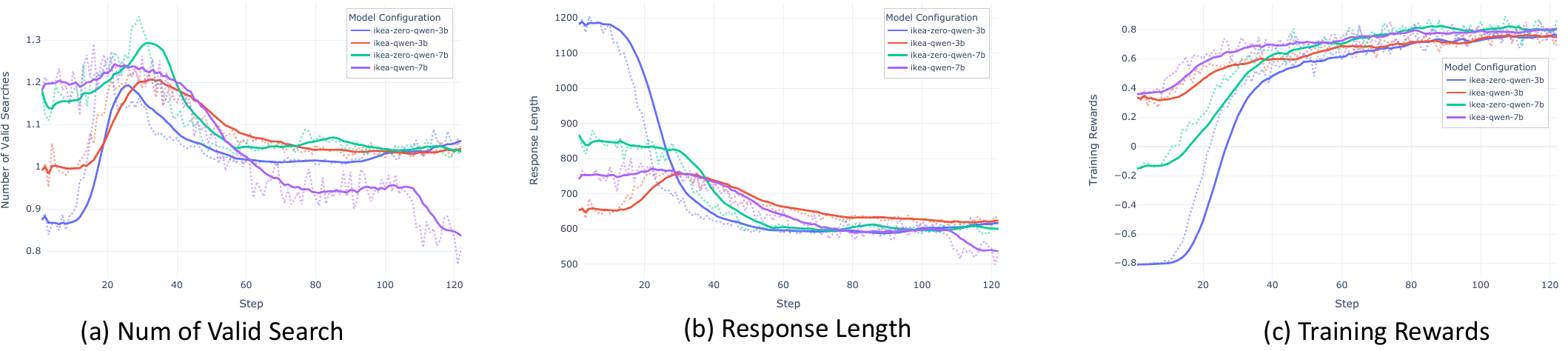}
    \caption{The training log of IKEA-3B-Zero, IKEA-3B, IKEA-7B-Zero and IKEA-7B. We show the curve of  number of valid searches, response length and trainign rewards.}
    \label{fig:main_training_log}
\end{figure*}

Experimental results are presented in Table \ref{tab:main_table}, with corresponding training logs illustrated in Figure \ref{fig:main_training_log}. A detailed analysis was conducted to demonstrate the advantages of the proposed method and provide insights for future research. 
It is posited that questions in the Easy subset primarily require knowledge within the knowledge boundary, whereas those in the Hard subset likely necessitate knowledge beyond it.

\paragraph{Baselines without parameter updates struggle to effectively synergize internal and external knowledge.}

"Direct" relies on internal knowledge, while "RAG" and "Iter-Retgen" (which performs iterative retrieval) use external knowledge. External knowledge significantly improves LLM performance on knowledge-intensive tasks, especially Hard subsets, indicating LLMs' internal knowledge deficiencies. However, constant retrieval causes conflicts and latency. Adaptive RAG methods, like IR-COT (LLM autonomously decides retrieval timing) and FLARE (retrieves based on low-probability tokens), aim to mitigate these issues. IR-COT improves performance on "Hard" tasks but degrades "Easy" ones due to knowledge conflicts. FLARE performs few retrievals, yielding performance similar to "Direct," suggesting token probability is not an effective retrieval trigger. The core finding is that internal and external knowledge must be synergistically used: internal when sufficient, external when insufficient. However, un-finetuned models struggle to autonomously determine when to leverage external knowledge.

\paragraph{Reinforcement learning baselines can effectively activate the model's capacity to solely utilize its internal knowledge or solely utilize external knowledge accessed via retrieval.}

R1, which relies on internal knowledge only based on reasoning, significantly improves performance on Easy subsets by reinforcing knowledge expressions through RL. However, its gains on Hard subsets are limited, highlighting the necessity of external knowledge retrieval. Search-R1, by generating search queries for external knowledge retrieval, addresses the internal knowledge deficit. It outperforms other methods (e.g., Iter-Retgen) with fewer retrievals, demonstrating that RL improves its planning and tool-using abilities for external knowledge access. While both R1 and Search-R1 show RL can boost the utilization of internal and external knowledge separately, neither method effectively integrates these two knowledge sources synergistically.

\paragraph{IKEA can adaptively combine internal and external knowledge for synergistic knowledge reasoning.}
During multiple rollouts, the model can choose to utilize only internal knowledge, only external knowledge, or a combination of both. Through a knowledge-boundary aware reward, RL encourages the model to leverage internal knowledge as much as possible when both internal and external knowledge are effective to reduce calls to retrieval tools, and to utilize retrieval to acquire external knowledge when internal knowledge is insufficient. As shown in the table, compared to R1, IKEA improves performance by over 10\%, with the improvement primarily coming from difficult subsets. This indicates that it can fully utilize external knowledge based on its internal knowledge. Compared to Search-R1, IKEA significantly reduces the number of retrievals while improving performance. This suggests that in the process of self-exploration, it learns to delineate its own knowledge boundaries, leveraging parametric knowledge as much as possible within these boundaries and retrieval knowledge outside of them. This not only effectively overcomes potential knowledge conflicts but also improves the overall process efficiency. It is worth noting that it also performs well on two out-of-distribution datasets, indicating that the knowledge-seeking behavior acquired through self-exploration can generalize effectively.

\paragraph{The IKEA training method is effective across models of different sizes and types.}

Figure \ref{fig:main_training_log} illustrates the IKEA training process based on different initial models. IKEA models, initialized from instruction-tuned models, start with higher rewards due to better instruction-following. IKEA-Zero models, starting from base models, begin with lower rewards but gradually learn the desired format. Both converge to similar reward levels by the end of training process, demonstrating that reinforcement learning can teach collaborative reasoning without cold start. Larger models (e.g., 7B vs. 3B) achieve higher initial and final rewards and converge faster. Retrieval counts initially increase before decreasing, indicating early benefit from more retrieval, followed by refinement to eliminate retrieval redundancy. Response length trends similarly for IKEA models (initial increase then decrease), while IKEA-Zero models show a consistent decrease, signifying the reduction of meaningless redundancy in the inital stage as they learn a fixed format.
\section{Ablation Study}

We conducted ablation studies based on Qwen2.5-3B-Instruct, which fully validated the effectiveness of the proposed method.

\subsection{The effects of reward design}
\begin{figure}[ht!]
    \centering
    \includegraphics[width=\linewidth]{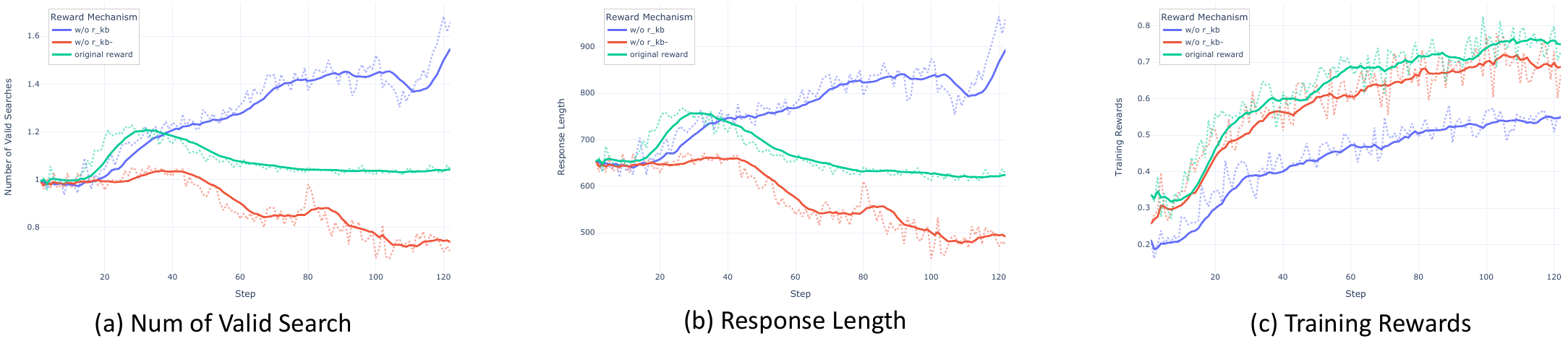}
    \caption{The training logs of different reward design. We show the curve of number of valid searches, response length and trainign rewards.}
    \label{fig:ablation_reward}
\end{figure}




\begin{table*}[ht!]
\centering
\caption{The ablation results of reward design.}
\label{tab:ablation_reward}
\resizebox{\textwidth}{!}{
\begin{tabular}{@{}lccccccccc@{}}
\toprule
\multirow{2}{*}{\textbf{Method}} & \multicolumn{2}{c}{\textbf{NQ}} & \multicolumn{2}{c}{\textbf{PopQA}} & \multicolumn{2}{c}{\textbf{HotpotQA}} & \multicolumn{2}{c}{\textbf{2Wiki}} & \multirow{2}{*}{\textbf{Avg}} \\
\cmidrule(lr){2-3} \cmidrule(lr){4-5} \cmidrule(lr){6-7} \cmidrule(lr){8-9}
& Easy & Hard & Easy & Hard & Easy & Hard & Easy & Hard & \\
\midrule

\rowcolor{lightpurple} IKEA (EM) & 72.46 & 31.44 & 79.69 & 33.59 & 69.92 & 20.11 & 59.37 & 20.70 & 48.41 \\
\rowcolor{lightpurple} \quad RT & 1.00 & 1.02 & 1.00 & 1.02 & 1.04 & 1.13 & 1.15 & 1.21 & 1.07 \\
IKEA w/o $r_{kb-}$ (EM) & 66.01 & 28.91 & 74.61 & 32.42 & 66.99 & 20.90 & 55.27 & 0.21 & 43.17 \\
\quad RT & 0.48 & 0.68 & 0.53 & 1.00 & 0.58 & 1.08 & 0.64 & 1.11 & 0.89 \\
IKEA w/o $r_{kb}$ (EM) & 71.09 & 34.57 & 76.37 & 32.23 & 70.12 & 25.59 & 53.32 & 25.20 & 48.56 \\
\quad RT & 1.40 & 1.54 & 1.35 & 1.63 & 1.94 & 2.12 & 2.40 & 2.48 & 1.86 \\

\bottomrule
\end{tabular}
} 
\end{table*}

We present the training process using different rewards in Figure \ref{fig:ablation_reward} and the final test results in Table \ref{tab:ablation_reward}.
Without the knowledge boundary aware reward ("$w/o$ $r_{kb}$"), both effective retrievals and response length show a consistent upward trend, significantly surpassing models with the original reward. This is because early in training, retrieval is more frequently rewarded than relying on parametric knowledge, leading to gradient updates that suppress the latter. Consequently, the model develops a bias for "retrieval > no retrieval", eventually maximizing reliance on retrieved knowledge, akin to the Search-R1 strategy.
For the "$w/o$ $r_{kb}$-" case (excluding the negative component of the knowledge boundary aware reward), retrieval count and response length are significantly less than the original reward. Because the positive reward component ($r_{kb}+$) encourages greater reliance on internal knowledge. This leads to incorrect generalization, where the model increasingly defaults to the R1 strategy even for questions requiring external knowledge.
Final results show that IKEA "$w/o$ $r_{kb}$" achieves a similar EM score but with significantly more retrievals. Conversely, IKEA "$w/o$ $r_{kb-}$" exhibits considerably degraded performance alongside a substantial decrease in retrievals.
Therefore, we conclude that an effective knowledge boundary aware reward function must appropriately balance internal and external knowledge utilization to achieve their synergistic application.

\subsection{The effects of dataset difficulty}
\begin{figure}[ht!]
    \centering
    \includegraphics[width=\linewidth]{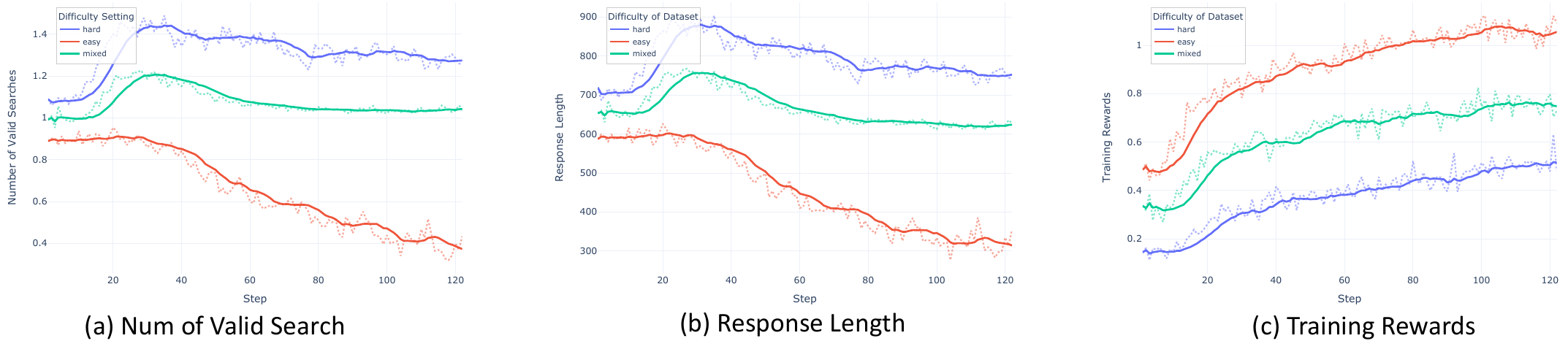}
    \caption{The training logs of different the difficulty of training datasets. We show the curve of number of valid searches, response length and trainign rewards.}
    \label{fig:ablation_dataset}
\end{figure}




\begin{table*}[ht!]
\centering
\caption{The ablation results of the difficulty of the training datasets.}
\label{tab:ablation_dataset}
\resizebox{\textwidth}{!}{
\begin{tabular}{@{}lccccccccc@{}}
\toprule
\multirow{2}{*}{\textbf{Method}} & \multicolumn{2}{c}{\textbf{NQ}} & \multicolumn{2}{c}{\textbf{PopQA}} & \multicolumn{2}{c}{\textbf{HotpotQA}} & \multicolumn{2}{c}{\textbf{2Wiki}} & \multirow{2}{*}{\textbf{Avg}} \\
\cmidrule(lr){2-3} \cmidrule(lr){4-5} \cmidrule(lr){6-7} \cmidrule(lr){8-9}
& Easy & Hard & Easy & Hard & Easy & Hard & Easy & Hard & \\
\midrule

\rowcolor{lightpurple} IKEA (EM) & 72.46 & 31.44 & 79.69 & 33.59 & 69.92 & 20.11 & 59.37 & 20.70 & 48.41 \\
\rowcolor{lightpurple} \quad RT & 1.00 & 1.02 & 1.00 & 1.02 & 1.04 & 1.13 & 1.15 & 1.21 & 1.07 \\
IKEA w/ easy (EM) & 66.99 & 21.88 & 76.17 & 25.59 & 66.70 & 15.43 & 56.45 & 16.80 & 43.25 \\
\quad RT & 0.28 & 0.54 & 0.29 & 0.80 & 0.34 & 0.84 & 0.16 & 0.70 & 0.49 \\
IKEA w/ hard (EM) & 66.02 & 33.98 & 75.39 & 0.35 & 64.65 & 25.00 & 46.09 & 23.63 & 41.89 \\
\quad RT & 1.03 & 1.07 & 1.05 & 1.11 & 1.46 & 1.59 & 2.08 & 2.10 & 1.44 \\

\bottomrule
\end{tabular}
} 
\end{table*}

We illustrate the training processes using datasets of varying difficulty in Figure \ref{fig:ablation_dataset} and present the final test results in Table \ref{tab:ablation_dataset}. 
Training on datasets of varying difficulty (easy, mixed, hard) revealed a consistent trend during training: Hard > Mixed > Easy for both effective number of searches and response length. This is because the model uses parametric knowledge for problems within its knowledge boundary and retrieval knowledge for those beyond it. Training on the Easy dataset showed a continuous decrease in retrieval attempts and response length, indicating that models converge to behaviors characteristic of the training data's difficulty. On the test set, both easy and hard variants of the IKEA model showed substantially lower Exact Match (EM) scores compared to the original. Retrieval attempts dropped significantly for the easy variant and increased substantially for the hard variant. This highlights that disproportionately favoring one type of knowledge hinders full performance, underscoring the importance of synergistically using both internal (parametric) and external (retrieval-based) knowledge for effective reasoning.


\section{Related Work}

\paragraph{RL for LLM-based Agent}
Reinforcement Learning (RL) \cite{wang2025reinforcementlearningenhancedllms} is a crucial technique for post-training Large Language Models (LLMs), enabling the alignment of pre-trained models' values \cite{ouyang2022traininglanguagemodelsfollow} and enhancing their capabilities in specific downstream tasks \cite{goldie2025syntheticdatageneration}. The community has developed various distinctive RL algorithms, such as PPO \cite{schulman2017proximalpolicyoptimizationalgorithms}, DPO \cite{rafailov2023direct}, RLOO \cite{ahmadian2024basicsrevisitingreinforcestyle}, ReMax \cite{li2024remaxsimpleeffectiveefficient}, and GRPO \cite{shao2024deepseekmathpushinglimitsmathematical}. Building upon this, by constructing different environments and reward functions, LLMs can be trained into intelligent agents capable of autonomous decision-making and interaction with the environment. A typical application in this area is the Search Agent \cite{jin2025searchr1trainingllmsreason, chen2025researchlearningreasonsearch, song2025r1searcherincentivizingsearchcapability}, which interacts with search engines to continuously acquire knowledge from the environment and perform reasoning, ultimately completing knowledge-intensive tasks.

\paragraph{The Knowledge Boundary of LLM}
Large Language Models (LLMs) possess parametric (internal) knowledge \cite{zheng2024reliablellmsknowledgebases} and can access external knowledge. The concept of \textit{Knowledge Boundary} \cite{li2024knowledgeboundarylargelanguage, xu2024rejection} distinguishes between these. This boundary is probed using template-based methods (evaluating responses to specific prompts \cite{petroni-etal-2019-language}) or internal state-based methods (classifying based on model features like hidden states \cite{chen2024teachinglargelanguagemodels} or SAEs \cite{zhao-etal-2025-steering}). Understanding this boundary is crucial for Retrieval Augmented Generation (RAG) models \cite{ren2024investigatingfactualknowledgeboundary} to adapt their behavior to different questions and avoid hallucinations.

\section{Conclusion and Limitations}
\label{sec:conclusion}

This paper introduced the Reinforced Internal-External Knowledge Synergistic Reasoning Agent (IKEA), an innovative approach to developing efficient and adaptive search agents. IKEA addresses critical limitations in existing RL-based search agents, namely the underutilization of internal knowledge, which can lead to redundant retrievals, potential knowledge conflicts, and increased inference latency. The core of IKEA lies in its ability to discern its own knowledge boundary, prioritizing the use of its internal parametric knowledge and resorting to external search only when the internal knowledge is deemed insufficient or uncertain. This is achieved through a novel knowledge-boundary aware reward function and a meticulously constructed knowledge-boundary aware training dataset. This approach significantly enhances reasoning efficiency and accuracy on knowledge-intensive tasks. Despite these achievements, IKEA's reliance on specific dataset construction and model probing for knowledge boundary awareness may limit its universal applicability, the reward function parameters might require grid searching, and the RL training process is computationally expensive. Future work could explore more dynamic knowledge boundary learning methods, investigate applicability across a broader range of tasks, and aim to reduce training resource requirements.

\bibliographystyle{plain}
\bibliography{custom}

\begin{thebibliography}{10}

\bibitem{ahmadian2024basicsrevisitingreinforcestyle}
Arash Ahmadian, Chris Cremer, Matthias Gallé, Marzieh Fadaee, Julia Kreutzer, Olivier Pietquin, Ahmet Üstün, and Sara Hooker.
\newblock Back to basics: Revisiting reinforce style optimization for learning from human feedback in llms, 2024.

\bibitem{chen2024teachinglargelanguagemodels}
Lida Chen, Zujie Liang, Xintao Wang, Jiaqing Liang, Yanghua Xiao, Feng Wei, Jinglei Chen, Zhenghong Hao, Bing Han, and Wei Wang.
\newblock Teaching large language models to express knowledge boundary from their own signals, 2024.

\bibitem{chen2025researchlearningreasonsearch}
Mingyang Chen, Tianpeng Li, Haoze Sun, Yijie Zhou, Chenzheng Zhu, Haofen Wang, Jeff~Z. Pan, Wen Zhang, Huajun Chen, Fan Yang, Zenan Zhou, and Weipeng Chen.
\newblock Research: Learning to reason with search for llms via reinforcement learning, 2025.

\bibitem{cheng2024understandinginterplayparametriccontextual}
Sitao Cheng, Liangming Pan, Xunjian Yin, Xinyi Wang, and William~Yang Wang.
\newblock Understanding the interplay between parametric and contextual knowledge for large language models, 2024.

\bibitem{deepseekai2025deepseekr1incentivizingreasoningcapability}
DeepSeek-AI, Daya Guo, Dejian Yang, Haowei Zhang, Junxiao Song, Ruoyu Zhang, Runxin Xu, Qihao Zhu, Shirong Ma, Peiyi Wang, Xiao Bi, Xiaokang Zhang, Xingkai Yu, Yu~Wu, Z.~F. Wu, Zhibin Gou, Zhihong Shao, Zhuoshu Li, Ziyi Gao, Aixin Liu, Bing Xue, Bingxuan Wang, Bochao Wu, Bei Feng, Chengda Lu, Chenggang Zhao, Chengqi Deng, Chenyu Zhang, Chong Ruan, Damai Dai, Deli Chen, Dongjie Ji, Erhang Li, Fangyun Lin, Fucong Dai, Fuli Luo, Guangbo Hao, Guanting Chen, Guowei Li, H.~Zhang, Han Bao, Hanwei Xu, Haocheng Wang, Honghui Ding, Huajian Xin, Huazuo Gao, Hui Qu, Hui Li, Jianzhong Guo, Jiashi Li, Jiawei Wang, Jingchang Chen, Jingyang Yuan, Junjie Qiu, Junlong Li, J.~L. Cai, Jiaqi Ni, Jian Liang, Jin Chen, Kai Dong, Kai Hu, Kaige Gao, Kang Guan, Kexin Huang, Kuai Yu, Lean Wang, Lecong Zhang, Liang Zhao, Litong Wang, Liyue Zhang, Lei Xu, Leyi Xia, Mingchuan Zhang, Minghua Zhang, Minghui Tang, Meng Li, Miaojun Wang, Mingming Li, Ning Tian, Panpan Huang, Peng Zhang, Qiancheng Wang, Qinyu Chen, Qiushi Du, Ruiqi Ge, Ruisong
  Zhang, Ruizhe Pan, Runji Wang, R.~J. Chen, R.~L. Jin, Ruyi Chen, Shanghao Lu, Shangyan Zhou, Shanhuang Chen, Shengfeng Ye, Shiyu Wang, Shuiping Yu, Shunfeng Zhou, Shuting Pan, S.~S. Li, Shuang Zhou, Shaoqing Wu, Shengfeng Ye, Tao Yun, Tian Pei, Tianyu Sun, T.~Wang, Wangding Zeng, Wanjia Zhao, Wen Liu, Wenfeng Liang, Wenjun Gao, Wenqin Yu, Wentao Zhang, W.~L. Xiao, Wei An, Xiaodong Liu, Xiaohan Wang, Xiaokang Chen, Xiaotao Nie, Xin Cheng, Xin Liu, Xin Xie, Xingchao Liu, Xinyu Yang, Xinyuan Li, Xuecheng Su, Xuheng Lin, X.~Q. Li, Xiangyue Jin, Xiaojin Shen, Xiaosha Chen, Xiaowen Sun, Xiaoxiang Wang, Xinnan Song, Xinyi Zhou, Xianzu Wang, Xinxia Shan, Y.~K. Li, Y.~Q. Wang, Y.~X. Wei, Yang Zhang, Yanhong Xu, Yao Li, Yao Zhao, Yaofeng Sun, Yaohui Wang, Yi~Yu, Yichao Zhang, Yifan Shi, Yiliang Xiong, Ying He, Yishi Piao, Yisong Wang, Yixuan Tan, Yiyang Ma, Yiyuan Liu, Yongqiang Guo, Yuan Ou, Yuduan Wang, Yue Gong, Yuheng Zou, Yujia He, Yunfan Xiong, Yuxiang Luo, Yuxiang You, Yuxuan Liu, Yuyang Zhou, Y.~X. Zhu,
  Yanhong Xu, Yanping Huang, Yaohui Li, Yi~Zheng, Yuchen Zhu, Yunxian Ma, Ying Tang, Yukun Zha, Yuting Yan, Z.~Z. Ren, Zehui Ren, Zhangli Sha, Zhe Fu, Zhean Xu, Zhenda Xie, Zhengyan Zhang, Zhewen Hao, Zhicheng Ma, Zhigang Yan, Zhiyu Wu, Zihui Gu, Zijia Zhu, Zijun Liu, Zilin Li, Ziwei Xie, Ziyang Song, Zizheng Pan, Zhen Huang, Zhipeng Xu, Zhongyu Zhang, and Zhen Zhang.
\newblock Deepseek-r1: Incentivizing reasoning capability in llms via reinforcement learning, 2025.

\bibitem{10.1145/3696410.3714717}
Guanting Dong, Yutao Zhu, Chenghao Zhang, Zechen Wang, Ji-Rong Wen, and Zhicheng Dou.
\newblock Understand what llm needs: Dual preference alignment for retrieval-augmented generation.
\newblock In {\em Proceedings of the ACM on Web Conference 2025}, WWW '25, page 4206–4225, New York, NY, USA, 2025. Association for Computing Machinery.

\bibitem{fang-etal-2024-enhancing}
Feiteng Fang, Yuelin Bai, Shiwen Ni, Min Yang, Xiaojun Chen, and Ruifeng Xu.
\newblock Enhancing noise robustness of retrieval-augmented language models with adaptive adversarial training.
\newblock In Lun-Wei Ku, Andre Martins, and Vivek Srikumar, editors, {\em Proceedings of the 62nd Annual Meeting of the Association for Computational Linguistics (Volume 1: Long Papers)}, pages 10028--10039, Bangkok, Thailand, August 2024. Association for Computational Linguistics.

\bibitem{gao2024retrievalaugmentedgenerationlargelanguage}
Yunfan Gao, Yun Xiong, Xinyu Gao, Kangxiang Jia, Jinliu Pan, Yuxi Bi, Yi~Dai, Jiawei Sun, Meng Wang, and Haofen Wang.
\newblock Retrieval-augmented generation for large language models: A survey, 2024.

\bibitem{goldie2025syntheticdatageneration}
Anna Goldie, Azalia Mirhoseini, Hao Zhou, Irene Cai, and Christopher~D. Manning.
\newblock Synthetic data generation \& multi-step rl for reasoning \& tool use, 2025.

\bibitem{guan2025deepragthinkingretrievalstep}
Xinyan Guan, Jiali Zeng, Fandong Meng, Chunlei Xin, Yaojie Lu, Hongyu Lin, Xianpei Han, Le~Sun, and Jie Zhou.
\newblock Deeprag: Thinking to retrieval step by step for large language models, 2025.

\bibitem{heinzerling-inui-2021-language}
Benjamin Heinzerling and Kentaro Inui.
\newblock Language models as knowledge bases: On entity representations, storage capacity, and paraphrased queries.
\newblock In Paola Merlo, Jorg Tiedemann, and Reut Tsarfaty, editors, {\em Proceedings of the 16th Conference of the European Chapter of the Association for Computational Linguistics: Main Volume}, pages 1772--1791, Online, April 2021. Association for Computational Linguistics.

\bibitem{ho-etal-2020-constructing}
Xanh Ho, Anh-Khoa Duong~Nguyen, Saku Sugawara, and Akiko Aizawa.
\newblock Constructing a multi-hop {QA} dataset for comprehensive evaluation of reasoning steps.
\newblock In Donia Scott, Nuria Bel, and Chengqing Zong, editors, {\em Proceedings of the 28th International Conference on Computational Linguistics}, pages 6609--6625, Barcelona, Spain (Online), December 2020. International Committee on Computational Linguistics.

\bibitem{Huang_2025}
Lei Huang, Weijiang Yu, Weitao Ma, Weihong Zhong, Zhangyin Feng, Haotian Wang, Qianglong Chen, Weihua Peng, Xiaocheng Feng, Bing Qin, and Ting Liu.
\newblock A survey on hallucination in large language models: Principles, taxonomy, challenges, and open questions.
\newblock {\em ACM Transactions on Information Systems}, 43(2):1–55, January 2025.

\bibitem{jeong2024adaptiveraglearningadaptretrievalaugmented}
Soyeong Jeong, Jinheon Baek, Sukmin Cho, Sung~Ju Hwang, and Jong~C. Park.
\newblock Adaptive-rag: Learning to adapt retrieval-augmented large language models through question complexity, 2024.

\bibitem{jiang2023activeretrievalaugmentedgeneration}
Zhengbao Jiang, Frank~F. Xu, Luyu Gao, Zhiqing Sun, Qian Liu, Jane Dwivedi-Yu, Yiming Yang, Jamie Callan, and Graham Neubig.
\newblock Active retrieval augmented generation, 2023.

\bibitem{jin2025searchr1trainingllmsreason}
Bowen Jin, Hansi Zeng, Zhenrui Yue, Jinsung Yoon, Sercan Arik, Dong Wang, Hamed Zamani, and Jiawei Han.
\newblock Search-r1: Training llms to reason and leverage search engines with reinforcement learning, 2025.

\bibitem{FlashRAG}
Jiajie Jin, Yutao Zhu, Xinyu Yang, Chenghao Zhang, and Zhicheng Dou.
\newblock Flashrag: {A} modular toolkit for efficient retrieval-augmented generation research.
\newblock {\em CoRR}, abs/2405.13576, 2024.

\bibitem{jin-etal-2024-cutting}
Zhuoran Jin, Pengfei Cao, Hongbang Yuan, Yubo Chen, Jiexin Xu, Huaijun Li, Xiaojian Jiang, Kang Liu, and Jun Zhao.
\newblock Cutting off the head ends the conflict: A mechanism for interpreting and mitigating knowledge conflicts in language models.
\newblock In Lun-Wei Ku, Andre Martins, and Vivek Srikumar, editors, {\em Findings of the Association for Computational Linguistics: ACL 2024}, pages 1193--1215, Bangkok, Thailand, August 2024. Association for Computational Linguistics.

\bibitem{kwiatkowski-etal-2019-natural}
Tom Kwiatkowski, Jennimaria Palomaki, Olivia Redfield, Michael Collins, Ankur Parikh, Chris Alberti, Danielle Epstein, Illia Polosukhin, Jacob Devlin, Kenton Lee, Kristina Toutanova, Llion Jones, Matthew Kelcey, Ming-Wei Chang, Andrew~M. Dai, Jakob Uszkoreit, Quoc Le, and Slav Petrov.
\newblock Natural questions: A benchmark for question answering research.
\newblock {\em Transactions of the Association for Computational Linguistics}, 7:452--466, 2019.

\bibitem{li2024knowledgeboundarylargelanguage}
Moxin Li, Yong Zhao, Yang Deng, Wenxuan Zhang, Shuaiyi Li, Wenya Xie, See-Kiong Ng, and Tat-Seng Chua.
\newblock Knowledge boundary of large language models: A survey, 2024.

\bibitem{li2024remaxsimpleeffectiveefficient}
Ziniu Li, Tian Xu, Yushun Zhang, Zhihang Lin, Yang Yu, Ruoyu Sun, and Zhi-Quan Luo.
\newblock Remax: A simple, effective, and efficient reinforcement learning method for aligning large language models, 2024.

\bibitem{mallen-etal-2023-trust}
Alex Mallen, Akari Asai, Victor Zhong, Rajarshi Das, Daniel Khashabi, and Hannaneh Hajishirzi.
\newblock When not to trust language models: Investigating effectiveness of parametric and non-parametric memories.
\newblock In Anna Rogers, Jordan Boyd-Graber, and Naoaki Okazaki, editors, {\em Proceedings of the 61st Annual Meeting of the Association for Computational Linguistics (Volume 1: Long Papers)}, pages 9802--9822, Toronto, Canada, July 2023. Association for Computational Linguistics.

\bibitem{mao-etal-2021-generation}
Yuning Mao, Pengcheng He, Xiaodong Liu, Yelong Shen, Jianfeng Gao, Jiawei Han, and Weizhu Chen.
\newblock Generation-augmented retrieval for open-domain question answering.
\newblock In Chengqing Zong, Fei Xia, Wenjie Li, and Roberto Navigli, editors, {\em Proceedings of the 59th Annual Meeting of the Association for Computational Linguistics and the 11th International Joint Conference on Natural Language Processing (Volume 1: Long Papers)}, pages 4089--4100, Online, August 2021. Association for Computational Linguistics.

\bibitem{ouyang2022traininglanguagemodelsfollow}
Long Ouyang, Jeff Wu, Xu~Jiang, Diogo Almeida, Carroll~L. Wainwright, Pamela Mishkin, Chong Zhang, Sandhini Agarwal, Katarina Slama, Alex Ray, John Schulman, Jacob Hilton, Fraser Kelton, Luke Miller, Maddie Simens, Amanda Askell, Peter Welinder, Paul Christiano, Jan Leike, and Ryan Lowe.
\newblock Training language models to follow instructions with human feedback, 2022.

\bibitem{petroni-etal-2019-language}
Fabio Petroni, Tim Rockt{\"a}schel, Sebastian Riedel, Patrick Lewis, Anton Bakhtin, Yuxiang Wu, and Alexander Miller.
\newblock Language models as knowledge bases?
\newblock In Kentaro Inui, Jing Jiang, Vincent Ng, and Xiaojun Wan, editors, {\em Proceedings of the 2019 Conference on Empirical Methods in Natural Language Processing and the 9th International Joint Conference on Natural Language Processing (EMNLP-IJCNLP)}, pages 2463--2473, Hong Kong, China, November 2019. Association for Computational Linguistics.

\bibitem{rafailov2023direct}
Rafael Rafailov, Archit Sharma, Eric Mitchell, Christopher~D Manning, Stefano Ermon, and Chelsea Finn.
\newblock Direct preference optimization: Your language model is secretly a reward model.
\newblock In {\em Thirty-seventh Conference on Neural Information Processing Systems}, 2023.

\bibitem{ren2024investigatingfactualknowledgeboundary}
Ruiyang Ren, Yuhao Wang, Yingqi Qu, Wayne~Xin Zhao, Jing Liu, Hao Tian, Hua Wu, Ji-Rong Wen, and Haifeng Wang.
\newblock Investigating the factual knowledge boundary of large language models with retrieval augmentation, 2024.

\bibitem{schulman2018highdimensionalcontinuouscontrolusing}
John Schulman, Philipp Moritz, Sergey Levine, Michael Jordan, and Pieter Abbeel.
\newblock High-dimensional continuous control using generalized advantage estimation, 2018.

\bibitem{schulman2017proximalpolicyoptimizationalgorithms}
John Schulman, Filip Wolski, Prafulla Dhariwal, Alec Radford, and Oleg Klimov.
\newblock Proximal policy optimization algorithms, 2017.

\bibitem{shao-etal-2023-enhancing}
Zhihong Shao, Yeyun Gong, Yelong Shen, Minlie Huang, Nan Duan, and Weizhu Chen.
\newblock Enhancing retrieval-augmented large language models with iterative retrieval-generation synergy.
\newblock In Houda Bouamor, Juan Pino, and Kalika Bali, editors, {\em Findings of the Association for Computational Linguistics: EMNLP 2023}, pages 9248--9274, Singapore, December 2023. Association for Computational Linguistics.

\bibitem{shao2024deepseekmathpushinglimitsmathematical}
Zhihong Shao, Peiyi Wang, Qihao Zhu, Runxin Xu, Junxiao Song, Xiao Bi, Haowei Zhang, Mingchuan Zhang, Y.~K. Li, Y.~Wu, and Daya Guo.
\newblock Deepseekmath: Pushing the limits of mathematical reasoning in open language models, 2024.

\bibitem{Sheng_2025}
Guangming Sheng, Chi Zhang, Zilingfeng Ye, Xibin Wu, Wang Zhang, Ru~Zhang, Yanghua Peng, Haibin Lin, and Chuan Wu.
\newblock Hybridflow: A flexible and efficient rlhf framework.
\newblock In {\em Proceedings of the Twentieth European Conference on Computer Systems}, EuroSys ’25, page 1279–1297. ACM, March 2025.

\bibitem{song2025r1searcherincentivizingsearchcapability}
Huatong Song, Jinhao Jiang, Yingqian Min, Jie Chen, Zhipeng Chen, Wayne~Xin Zhao, Lei Fang, and Ji-Rong Wen.
\newblock R1-searcher: Incentivizing the search capability in llms via reinforcement learning, 2025.

\bibitem{su2025crossingrewardbridgeexpanding}
Yi~Su, Dian Yu, Linfeng Song, Juntao Li, Haitao Mi, Zhaopeng Tu, Min Zhang, and Dong Yu.
\newblock Crossing the reward bridge: Expanding rl with verifiable rewards across diverse domains, 2025.

\bibitem{trivedi-etal-2023-interleaving}
Harsh Trivedi, Niranjan Balasubramanian, Tushar Khot, and Ashish Sabharwal.
\newblock Interleaving retrieval with chain-of-thought reasoning for knowledge-intensive multi-step questions.
\newblock In Anna Rogers, Jordan Boyd-Graber, and Naoaki Okazaki, editors, {\em Proceedings of the 61st Annual Meeting of the Association for Computational Linguistics (Volume 1: Long Papers)}, pages 10014--10037, Toronto, Canada, July 2023. Association for Computational Linguistics.

\bibitem{wang2025chainofretrievalaugmentedgeneration}
Liang Wang, Haonan Chen, Nan Yang, Xiaolong Huang, Zhicheng Dou, and Furu Wei.
\newblock Chain-of-retrieval augmented generation, 2025.

\bibitem{wang2022text}
Liang Wang, Nan Yang, Xiaolong Huang, Binxing Jiao, Linjun Yang, Daxin Jiang, Rangan Majumder, and Furu Wei.
\newblock Text embeddings by weakly-supervised contrastive pre-training.
\newblock {\em arXiv preprint arXiv:2212.03533}, 2022.

\bibitem{wang2025reinforcementlearningenhancedllms}
Shuhe Wang, Shengyu Zhang, Jie Zhang, Runyi Hu, Xiaoya Li, Tianwei Zhang, Jiwei Li, Fei Wu, Guoyin Wang, and Eduard Hovy.
\newblock Reinforcement learning enhanced llms: A survey, 2025.

\bibitem{wen2024perception}
Zhihua Wen, Zhiliang Tian, Zexin Jian, Zhen Huang, Pei Ke, Yifu Gao, Minlie Huang, and Dongsheng Li.
\newblock Perception of knowledge boundary for large language models through semi-open-ended question answering.
\newblock In {\em The Thirty-eighth Annual Conference on Neural Information Processing Systems}, 2024.

\bibitem{xu2024rejection}
Hongshen Xu, Zichen Zhu, Situo Zhang, Da~Ma, Shuai Fan, Lu~Chen, and Kai Yu.
\newblock Rejection improves reliability: Training {LLM}s to refuse unknown questions using {RL} from knowledge feedback.
\newblock In {\em First Conference on Language Modeling}, 2024.

\bibitem{xu-etal-2024-knowledge-conflicts}
Rongwu Xu, Zehan Qi, Zhijiang Guo, Cunxiang Wang, Hongru Wang, Yue Zhang, and Wei Xu.
\newblock Knowledge conflicts for {LLM}s: A survey.
\newblock In Yaser Al-Onaizan, Mohit Bansal, and Yun-Nung Chen, editors, {\em Proceedings of the 2024 Conference on Empirical Methods in Natural Language Processing}, pages 8541--8565, Miami, Florida, USA, November 2024. Association for Computational Linguistics.

\bibitem{yang-etal-2018-hotpotqa}
Zhilin Yang, Peng Qi, Saizheng Zhang, Yoshua Bengio, William Cohen, Ruslan Salakhutdinov, and Christopher~D. Manning.
\newblock {H}otpot{QA}: A dataset for diverse, explainable multi-hop question answering.
\newblock In Ellen Riloff, David Chiang, Julia Hockenmaier, and Jun{'}ichi Tsujii, editors, {\em Proceedings of the 2018 Conference on Empirical Methods in Natural Language Processing}, pages 2369--2380, Brussels, Belgium, October-November 2018. Association for Computational Linguistics.

\bibitem{yu2024autoragautonomousretrievalaugmentedgeneration}
Tian Yu, Shaolei Zhang, and Yang Feng.
\newblock Auto-rag: Autonomous retrieval-augmented generation for large language models, 2024.

\bibitem{zhao-etal-2025-steering}
Yu~Zhao, Alessio Devoto, Giwon Hong, Xiaotang Du, Aryo~Pradipta Gema, Hongru Wang, Xuanli He, Kam-Fai Wong, and Pasquale Minervini.
\newblock Steering knowledge selection behaviours in {LLM}s via {SAE}-based representation engineering.
\newblock In Luis Chiruzzo, Alan Ritter, and Lu~Wang, editors, {\em Proceedings of the 2025 Conference of the Nations of the Americas Chapter of the Association for Computational Linguistics: Human Language Technologies (Volume 1: Long Papers)}, pages 5117--5136, Albuquerque, New Mexico, April 2025. Association for Computational Linguistics.

\bibitem{zhao2024knowingllmsknowsimple}
Yukun Zhao, Lingyong Yan, Weiwei Sun, Guoliang Xing, Chong Meng, Shuaiqiang Wang, Zhicong Cheng, Zhaochun Ren, and Dawei Yin.
\newblock Knowing what llms do not know: A simple yet effective self-detection method, 2024.

\bibitem{zheng2024reliablellmsknowledgebases}
Danna Zheng, Mirella Lapata, and Jeff~Z. Pan.
\newblock How reliable are llms as knowledge bases? re-thinking facutality and consistency, 2024.

\end{thebibliography}

\newpage
\appendix
\section{IKEA agent template}
\label{app:prompt}
We use the system template in Table \ref{tab:prompt} to prompt the agent to interact with the environment:

\begin{table*}[htbp]
\centering
\begin{tabularx}{\textwidth}{X}
\toprule 
You are an expert assistant capable of solving knowledge-intensive tasks efficiently. You will be given a question to answer as accurately as possible. 

You can use your own knowledge or call external search engines to gather additional information, but searching should only occur when necessary. Specifically, you should search only when encountering a clear knowledge gap or uncertainty that prevents you from confidently answering the question. 

To arrive at the answer, you will proceed step-by-step in a structured cycle of '<think>thinking content</think>', '<search>search query</search>' (optional), and '<context>returned external information</context>' (optional) sequences. You can only generate content within these special tags.

Remember that <search>xxx</search> and <context>xxx</context> are optional. You can skip them if you have enough knowledge to answer the question. And skip is them is encouraged and preferable.

Thinking Phase (<think>): Recall your own knowledge, analyze current information, and decide whether further search is needed. If enough knowledge is available, skip searching. For question, it may be decomposed into sub-questions for you to think about. Some sub-questions may be answered by searching, while others may not. You can also use the <think> tag to express your uncertainty about the sub-question.

Searching Phase (<search>): Formulate a search query only if required to fill a knowledge gap or verify uncertainty. Skip if unnecessary.
Information Phase (<context>): Use search results as context for further steps. If no search was performed, proceed without this phase.

Answering Phase (<answer>): Provide a concise and accurate answer within <answer> tags once you have enough knowledge. The answer should be short and precise, such as <answer> Beijing </answer>.

Here are a few examples:

---

Example 1: search is needed, search more than once

Question: xxx

<think> xxx </think>

search> xxx </search>

<context> xxx </context>

<think> xxx </think>

(search more than once)

<think> xxx </think>

<answer> xx </answer>

Example 2: search is needed, only search once

Question: xxx?

<think> xxx </think>

<search> xxx </search>

<context> xxx </context>

<think> xxx </think>

<answer> xxx </answer>

---

Example 3: search is not needed

Question: xxx?

<think> xxx </think>

<answer> xxx </answer>

---

You can search 0 - N times. 0 is preferable. Each search should be focused on one sub-question.

The answer within <answer> tags should be short and precise, such as <answer> yes </answer>.

Now it is your turn to answer the question.

Question: \{question\}
\\
\bottomrule
\end{tabularx}
\caption{System prompt of IKEA.}
\label{tab:prompt}
\end{table*}

\section{Dataset Construction}
\label{app:dataset}
We use NQ \cite{kwiatkowski-etal-2019-natural} and HotpotQA \cite{ho-etal-2020-constructing} as the in-distribution datasets. We use the PopQA \cite{mallen-etal-2023-trust} and 2Wikimultihopqa \cite{ho-etal-2020-constructing} as the out-of-distribution datasets. Following the knowledge-boundary training dataset construction method, we construct easy and hard subset for each dataset. We use the Qwen-2.5-3B-Instruct as the sampling model. There are 512 examples in each subset of each dataset.

\section{Baselines}
\label{app:baseline}
We compared methods that do not require training (e.g., zero-shot or few-shot prompting), those that utilize Supervised Fine-Tuning (SFT) and Direct Preference Optimization (DPO), and reinforcement learning-based approaches. The corresponding baselines are shown as follows:
\begin{itemize}
    \item \textbf{Direct} We directly prompt the model to answer the relevant question using only its internal knowledge.
    \item \textbf{RAG} We retrieve documents using the question and prompt the model to answer the relevant question relying solely on the retrieved knowledge.
    \item \textbf{Iter-Retgen} \cite{shao-etal-2023-enhancing} It is an iterative retrieval-generation method that achieves strong performance by synergizing parametric and non-parametric knowledge. We set the default ret-gen turns as 4.
    \item  \textbf{IR-COT} \cite{trivedi-etal-2023-interleaving} It is method for multi-step question answering, which interleaves retrieval with steps in the chain-of-thought, using CoT to guide retrieval and retrieval results to improve CoT. It will adatpively determine the turns of retrieval according to the knowledge needs. And we set the max turns as 4.
    \item \textbf{FLARE} \cite{jiang2023activeretrievalaugmentedgeneration} This method introduces a forward-looking active retrieval-augmented generation (FLARE) approach that iteratively uses predictions of upcoming sentences to anticipate future content and retrieves relevant documents when a sentence contains low-confidence tokens, in order to regenerate that sentence. It uses a specific criteria to determine the retrieval timing. We set the max number of search as 4.
    \item \textbf{DeepRAG} \cite{guan2025deepragthinkingretrievalstep} This method introduces a framework that models retrieval-augmented generation as a Markov Decision Process (MDP), enabling strategic and adaptive retrieval to improve retrieval efficiency and answer accuracy. It collects offline trajectories to finetune the base model with SFT and DPO.
    \item \textbf{R1} \cite{deepseekai2025deepseekr1incentivizingreasoningcapability} It uses reinforcement learning to encourage the model to reason in order to activate its internal knowledge. This method only uses the internal knowledge.
    \item \textbf{Search-R1} \cite{jin2025searchr1trainingllmsreason, song2025r1searcherincentivizingsearchcapability, chen2025researchlearningreasonsearch} The model's capacity to employ external retrieval tools is activated via multi-turn reinforcement learning. This technique exclusively relies on the model's external knowledge.  We set the max number of search as 4.
\end{itemize}

\section{Implementation Details}
\label{app:implementation}
We use e5-base \cite{wang2022text} as the retriever model and wikipedia2018 as the corpus for retrieval. We employ Qwen2.5-3B(-Instruct) and Qwen2.5-7B(-Instruct) as the initial models. Models with the "-Zero" suffix are trained from the Base model, while those without it are trained from the Instruct model. we use FlashRAG \cite{FlashRAG} to reproduce the baseline results. We utilize the verl \cite{Sheng_2025} framework for training. GRPO \cite{shao2024deepseekmathpushinglimitsmathematical} is used as the reinforcement learning algorithm. We use the NQ and HotpotQA to construct training datasets. For each one, we sample 4000 easy samples and 4000 hard samples. We set the number of rollouts as 16 for one task. We set the learning rate as 5e-7, warmup ratio as 75\%, batch size as 256, training steps as 120. We set $r_{kb+}$ as 0.6 and $r_{kb-}$ as 0.05, $RT_{max}$ as 3. We use 8 A100 GPUs for all the experiments.

\newpage
\section*{NeurIPS Paper Checklist}

The checklist is designed to encourage best practices for responsible machine learning research, addressing issues of reproducibility, transparency, research ethics, and societal impact. Do not remove the checklist: {\bf The papers not including the checklist will be desk rejected.} The checklist should follow the references and follow the (optional) supplemental material.  The checklist does NOT count towards the page
limit. 

Please read the checklist guidelines carefully for information on how to answer these questions. For each question in the checklist:
\begin{itemize}
    \item You should answer \answerYes{}, \answerNo{}, or \answerNA{}.
    \item \answerNA{} means either that the question is Not Applicable for that particular paper or the relevant information is Not Available.
    \item Please provide a short (1–2 sentence) justification right after your answer (even for NA). 
\end{itemize}

{\bf The checklist answers are an integral part of your paper submission.} They are visible to the reviewers, area chairs, senior area chairs, and ethics reviewers. You will be asked to also include it (after eventual revisions) with the final version of your paper, and its final version will be published with the paper.

The reviewers of your paper will be asked to use the checklist as one of the factors in their evaluation. While "\answerYes{}" is generally preferable to "\answerNo{}", it is perfectly acceptable to answer "\answerNo{}" provided a proper justification is given (e.g., "error bars are not reported because it would be too computationally expensive" or "we were unable to find the license for the dataset we used"). In general, answering "\answerNo{}" or "\answerNA{}" is not grounds for rejection. While the questions are phrased in a binary way, we acknowledge that the true answer is often more nuanced, so please just use your best judgment and write a justification to elaborate. All supporting evidence can appear either in the main paper or the supplemental material, provided in appendix. If you answer \answerYes{} to a question, in the justification please point to the section(s) where related material for the question can be found.

IMPORTANT, please:
\begin{itemize}
    \item {\bf Delete this instruction block, but keep the section heading ``NeurIPS Paper Checklist"},
    \item  {\bf Keep the checklist subsection headings, questions/answers and guidelines below.}
    \item {\bf Do not modify the questions and only use the provided macros for your answers}.
\end{itemize}


\begin{enumerate}

\item {\bf Claims}
    \item[] Question: Do the main claims made in the abstract and introduction accurately reflect the paper's contributions and scope?
    \item[] Answer: \answerYes{} 
    \item[] Justification: We clarify the contributions and the scope in abstract and the introduction.

\item {\bf Limitations}
    \item[] Question: Does the paper discuss the limitations of the work performed by the authors?
    \item[] Answer: \answerYes{} 
    \item[] Justification: We discuss the limitations of the work in Section \ref{sec:conclusion}.

\item {\bf Theory assumptions and proofs}
    \item[] Question: For each theoretical result, does the paper provide the full set of assumptions and a complete (and correct) proof?
    \item[] Answer: \answerNA{} 
    \item[] Justification: The paper does not include theoretical results.

    \item {\bf Experimental result reproducibility}
    \item[] Question: Does the paper fully disclose all the information needed to reproduce the main experimental results of the paper to the extent that it affects the main claims and/or conclusions of the paper (regardless of whether the code and data are provided or not)?
    \item[] Answer: \answerYes{} 
    \item[] Justification: We provide the details in Section \ref{app:prompt}, \ref{app:dataset} and \ref{app:implementation}.

\item {\bf Open access to data and code}
    \item[] Question: Does the paper provide open access to the data and code, with sufficient instructions to faithfully reproduce the main experimental results, as described in supplemental material?
    \item[] Answer: \answerYes{} 
    \item[] Justification: We provide the data and code in supplementary materials.

\item {\bf Experimental setting/details}
    \item[] Question: Does the paper specify all the training and test details (e.g., data splits, hyperparameters, how they were chosen, type of optimizer, etc.) necessary to understand the results?
    \item[] Answer: \answerYes{} 
    \item[] Justification: We provide the details in Section \ref{app:prompt}, \ref{app:dataset} and \ref{app:implementation}.

\item {\bf Experiment statistical significance}
    \item[] Question: Does the paper report error bars suitably and correctly defined or other appropriate information about the statistical significance of the experiments?
    \item[] Answer: \answerNo{} 
    \item[] Justification: The computation resources are too expensive for our lab to repeat much more times.

\item {\bf Experiments compute resources}
    \item[] Question: For each experiment, does the paper provide sufficient information on the computer resources (type of compute workers, memory, time of execution) needed to reproduce the experiments?
    \item[] Answer: \answerYes{} 
    \item[] Justification: We show it in Section \ref{app:implementation}.
    
\item {\bf Code of ethics}
    \item[] Question: Does the research conducted in the paper conform, in every respect, with the NeurIPS Code of Ethics \url{https://neurips.cc/public/EthicsGuidelines}?
    \item[] Answer: \answerYes{} 
    \item[] Justification: This paper conforms, in every respect, with the NeurIPS Code of Ethics.

\item {\bf Broader impacts}
    \item[] Question: Does the paper discuss both potential positive societal impacts and negative societal impacts of the work performed?
    \item[] Answer: \answerNA{} 
    \item[] Justification: We think our work will not have a significant social impact.
    
\item {\bf Safeguards}
    \item[] Question: Does the paper describe safeguards that have been put in place for responsible release of data or models that have a high risk for misuse (e.g., pretrained language models, image generators, or scraped datasets)?
    \item[] Answer: \answerNA{} 
    \item[] Justification: The paper poses no such risks.

\item {\bf Licenses for existing assets}
    \item[] Question: Are the creators or original owners of assets (e.g., code, data, models), used in the paper, properly credited and are the license and terms of use explicitly mentioned and properly respected?
    \item[] Answer: \answerYes{} 
    \item[] Justification: All assets used in this paper are properly credited.

\item {\bf New assets}
    \item[] Question: Are new assets introduced in the paper well documented and is the documentation provided alongside the assets?
    \item[] Answer: \answerYes{} 
    \item[] Justification: We provide the documentation in the code repo.

\item {\bf Crowdsourcing and research with human subjects}
    \item[] Question: For crowdsourcing experiments and research with human subjects, does the paper include the full text of instructions given to participants and screenshots, if applicable, as well as details about compensation (if any)? 
    \item[] Answer: \answerNA{} 
    \item[] Justification: The paper does not involve crowdsourcing nor research with human subjects.

\item {\bf Institutional review board (IRB) approvals or equivalent for research with human subjects}
    \item[] Question: Does the paper describe potential risks incurred by study participants, whether such risks were disclosed to the subjects, and whether Institutional Review Board (IRB) approvals (or an equivalent approval/review based on the requirements of your country or institution) were obtained?
    \item[] Answer: \answerNA{} 
    \item[] Justification: The paper does not involve crowdsourcing nor research with human subjects.

\item {\bf Declaration of LLM usage}
    \item[] Question: Does the paper describe the usage of LLMs if it is an important, original, or non-standard component of the core methods in this research? Note that if the LLM is used only for writing, editing, or formatting purposes and does not impact the core methodology, scientific rigorousness, or originality of the research, declaration is not required.
    \item[] Answer: \answerNA{} 
    \item[] Justification: The core method development in this research does not involve LLMs as any important, original, or non-standard components.

\end{enumerate}

\end{document}